\documentclass[runningheads]{llncs}

\usepackage[year=2026,ID=14845]{eccv}

\usepackage{eccvabbrv}

\usepackage{graphicx}
\usepackage{booktabs}

\usepackage[accsupp]{axessibility}  

\usepackage{hyperref}

\usepackage{siunitx}

\usepackage{orcidlink}
\usepackage{multirow}

\begin{document}

\title{DRS-VPT: Directly Relocalizing in a Scan with Vision Point Transformers}


\author{Lanke Frank Tarimo Fu\orcidlink{0000-0002-5226-6439} \and
Maurice Fallon\orcidlink{0000-0003-2940-0879}}


\institute{University of Oxford, Oxford OX2 6NN, United Kingdom\\
\email{\{fu,mfallon\}@robots.ox.ac.uk}}

\maketitle

\begin{abstract}
We present \textbf{DRS-VPT}, a feed-forward transformer architecture for foundational image-to-scan registration.
Given query images and a reference 3D point cloud, the model predicts the scan pose and point map alongside the poses and point maps of each camera, all expressed in the first camera's frame.
It additionally predicts a coarse-to-fine pyramid of per-point and per-pixel features for direct reprojective alignment of the scan to the first image.
This formulation unifies downstream tasks such as camera--LiDAR calibration in autonomous driving and indoor camera-to-map relocalization.
A single DRS-VPT model achieves state-of-the-art performance for image-to-LiDAR registration in autonomous driving, competitive indoor relocalization without training map-specific weights, and strong zero-shot transfer to unseen environments.
We also show qualitatively that the model learns complex scan-to-image projection properties such as occlusion of back-facing points.
\end{abstract}

\begin{figure*}[t]
    \centering
    \includegraphics[width=0.9\textwidth]{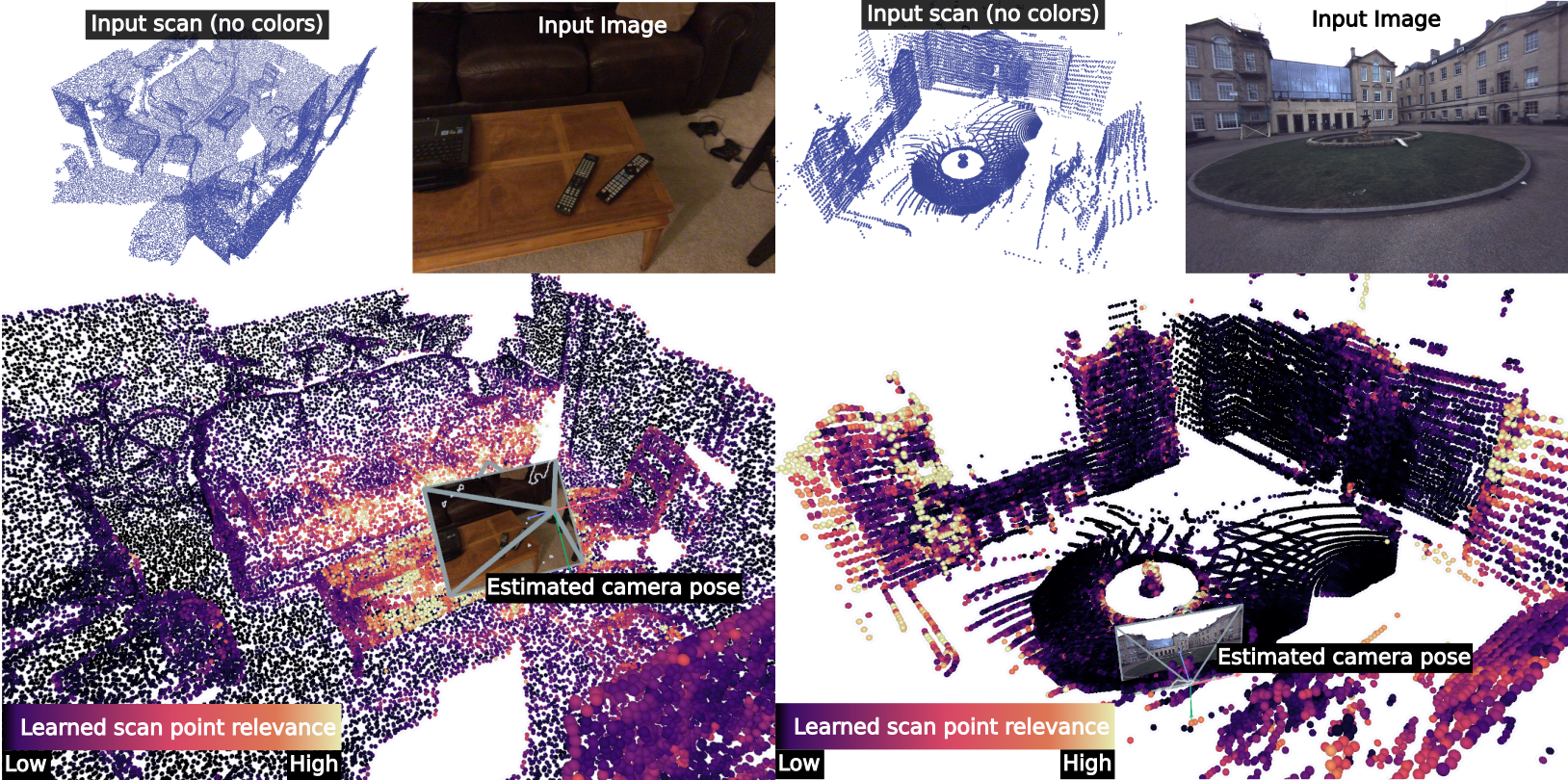}
    \caption{As input, our model ingests raw point clouds (without color information), alongside images taken from different poses.
    The model accurately registers the pose of the camera in the point cloud. (Left) camera to dense map localization on a 12scenes (unseen during training).
    (Right) camera to sparse scan localization on the Oxford Spires dataset.}
    \vspace{-20pt}
    \label{fig:task_overview}
\end{figure*}

\section{Introduction}
\label{sec:intro}

Recent progress in feed-forward visual geometry has shown that large
transformer architectures can infer dense scene structure and camera
poses directly from images, without requiring explicit calibration or
iterative optimization. Models such as DUSt3R~\cite{dust3r}, MASt3R~\cite{mast3r}, VGGT~\cite{wang2025vggt}, and MapAnything~\cite{mapanything} demonstrate that generalizable multi-view geometry can be learned at scale.
However, these approaches operate purely in the image domain. When geometry priors are available~\cite{mapanything, power_2025_CVPR}, they are assumed to be co-registered with an image view.
In many applications---including robotics, autonomous driving, and digital surveying---scene geometry is instead captured independently by sensors such as LiDAR or structured-light scanners, and is not aligned to any camera.
This is the problem of \emph{image-to-scan registration}: aligning a query image to a reference 3D point cloud (see~\autoref{fig:task_overview}).

Image-to-scan registration takes different forms. In autonomous driving, it arises as camera--LiDAR calibration. Classical methods~\cite{kaess_zhou, diffcal, levinson2013automatic} require controlled targets, known intrinsics, or accurate initialization; learned approaches~\cite{dxqnet, voxcal} assume small pose offsets and train on a single sensor configuration, limiting generalization.
Similarly, in indoor and outdoor camera-to-map relocalization, a query image must be registered against a 3D scan of the environment.
Existing methods~\cite{sarlin21pixloc,taira2018inloc,zhang2026visual} perform image-to-scan alignment by either lifting image features into SFM reconstructions, or rendering image views from dense XYZRGB maps where an expensive TLS sensor provides per-point color.
Both approaches rely on an external source of pre-aligned per-point color baked into the 3D map.

In this work, we introduce \textbf{DRS-VPT}, a feed-forward transformer architecture for image-to-scan registration.
By training at scale across diverse sensors and environments,
the model learns true cross-modal alignment between raw point clouds and images without baking image features into the input 3D points, or prior knowledge of the scan's pose.
For each camera, the model predicts a dense point map and camera pose.
For scan alignment, the model predicts a coarse scan pose and dense cross-modal feature pyramids for both the scan and first image, which together drive differentiable direct image-to-scan alignment, achieving state-of-the-art camera--LiDAR registration and competitive relocalization against raw point cloud maps.

Our contributions are as follows:
\vspace{-5pt}
\begin{itemize}
\item We introduce \textbf{DRS-VPT}, a transformer architecture that unifies feed forward geometry prediction with differentiable direct image-to-scan alignment, enabling image-to-scan registration without pose initialization or per-point image features in the reference map.

\item Through foundational training across diverse sensors and environments, a single DRS-VPT model achieves state-of-the-art camera--LiDAR registration and competitive relocalization against raw point cloud maps, generalizing across domains where existing single-dataset methods fail catastrophically.

\item As we show in an ablation, reprojective feature-metric alignment significantly outperforms geometric point-alignment alternatives such as ICP and Kabsch, reducing mean translation error by $5\times$ on camera--LiDAR registration.

\end{itemize}

\section{Related Work}

\paragraph{Learned Wide Baseline Image--Scan Alignment.} Motivated by online camera-LiDAR calibration fault recovery for autonomous driving, DeepI2P~\cite{Li2021DeepI2PIC} introduced image--scan alignment from wide offsets (${\approx}$\,5\,m) without pose initialization.
DeepI2P uses a decoupled two-stage design: a cross-modal classifier trained with binary cross-entropy labels each scan point inside or outside the camera frustum, then a separate inverse camera projection solver recovers the pose, with no gradient flowing back to the classifier.
Later, CorrI2P~\cite{corri2p}, RelaI2P~\cite{relai2p}, and VP2P~\cite{vp2p} also employ the two-stage design but instead supervise point-to-pixel correspondences end-to-end, by regressing pose via a differentiable PnP solver.
More recently, TrafficLoc~\cite{trafficloc} additionally applies attention supervision to improve cross-modal feature alignment.
Trained independently on KITTI and nuScenes, these methods generalize poorly across datasets---even between the two---a failure mode we confirm in our experiments.

\paragraph{Learned Image--Dense Scan Alignment.} A related line of work targets image-to-dense-point-cloud alignment under high overlap.
P2-Net~\cite{p2net} jointly detects and describes keypoints in a shared cross-modal descriptor space, without feature fusion between modalities.
2D3D-MATR~\cite{matr2d3d} removes the detection step, matching coarse-to-fine patch descriptors with a detection-free transformer that fuses image and point cloud features.
UniCorrn~\cite{unicorrn} unifies 2D--2D, 2D--3D, and 3D--3D correspondence under a single shared-weight transformer.
These methods address a considerably more constrained problem than the camera--LiDAR setting above.
The reference point cloud is back-projected from a dense depth map, and image--scan overlap exceeds $50\%$.
Neither of these conditions holds when a narrow-FOV camera registers against a sparse 360$^{\circ}$ LiDAR sweep.

\paragraph{Image--Scan Alignment in Camera Relocalization.} Motivated by camera relocalization against pre-built 3D maps, these methods register query images to 3D scans by embedding image appearance into the map.
PixLoc~\cite{sarlin21pixloc} lifts appearance features from mapping images into an SfM reconstruction and retrieves query poses via feature-metric direct alignment.
InLoc~\cite{taira2018inloc} and Zhang~\etal~\cite{zhang2026visual} instead render synthetic image views---from a posed RGBD database and a TLS XYZRGB map respectively---and match against the query image.
All require per-point image appearance baked into the map, precluding localization against geometry-only scans.

\paragraph{Fine Camera--LiDAR Alignment.} Accurate camera--LiDAR alignment is fundamental to sensor fusion between these modalities.
Where bespoke calibration targets are available, classical methods optimize the joint alignment of plane and line observations across both sensors~\cite{kaess_zhou}, or the alignment of planar geometry alongside the intensity patterns of the target~\cite{diffcal}.
In unstructured environments, calibration instead relies on mutual-information maximization~\cite{gaurav_pandey}, intensity--edge matching~\cite{levinson2013automatic}, or online alignment of LiDAR and camera intensity signals~\cite{pascoe_calib, napier}.
Learning-based approaches~\cite{lccnet, calibNet, RGKCNet} directly regress the camera--LiDAR transformation from paired observations, but exhibit poor generalization across sensor configurations~\cite{voxcal}.
DXQNet~\cite{dxqnet} and VoxCal~\cite{voxcal} improve on this by coupling differentiable pose optimization into the training loop, allowing pose supervision to shape the learned features.
All these methods assume a relatively accurate initial guess pose---they perform local refinement, not recovery from wide offsets.

\paragraph{Feed-Forward Visual Geometry Models.}Early convolutional models for joint depth and motion estimation~\cite{ummenhofer2017demon, deeptam} showed limited generalisation, prompting methods to embed geometry-grounded differentiable optimization into training~\cite{Teed2018DeepV2DVT, sfmlearner, droidslam, vggsfm}.
DUSt3R~\cite{dust3r} then demonstrated that dense two-view geometry can be solved end-to-end with a large transformer trained at scale, motivating MASt3R~\cite{mast3r} and VGGT~\cite{wang2025vggt}, which extended this to many views.
Pow3R~\cite{power_2025_CVPR} and MapAnything~\cite{mapanything} further extend these architectures to ingest ground-truth geometry alongside images, but require the geometry to be frame-aligned to camera views.


\begin{figure}[t]
    \vspace{-1mm}
    \centering
    \includegraphics[width=0.95\linewidth]{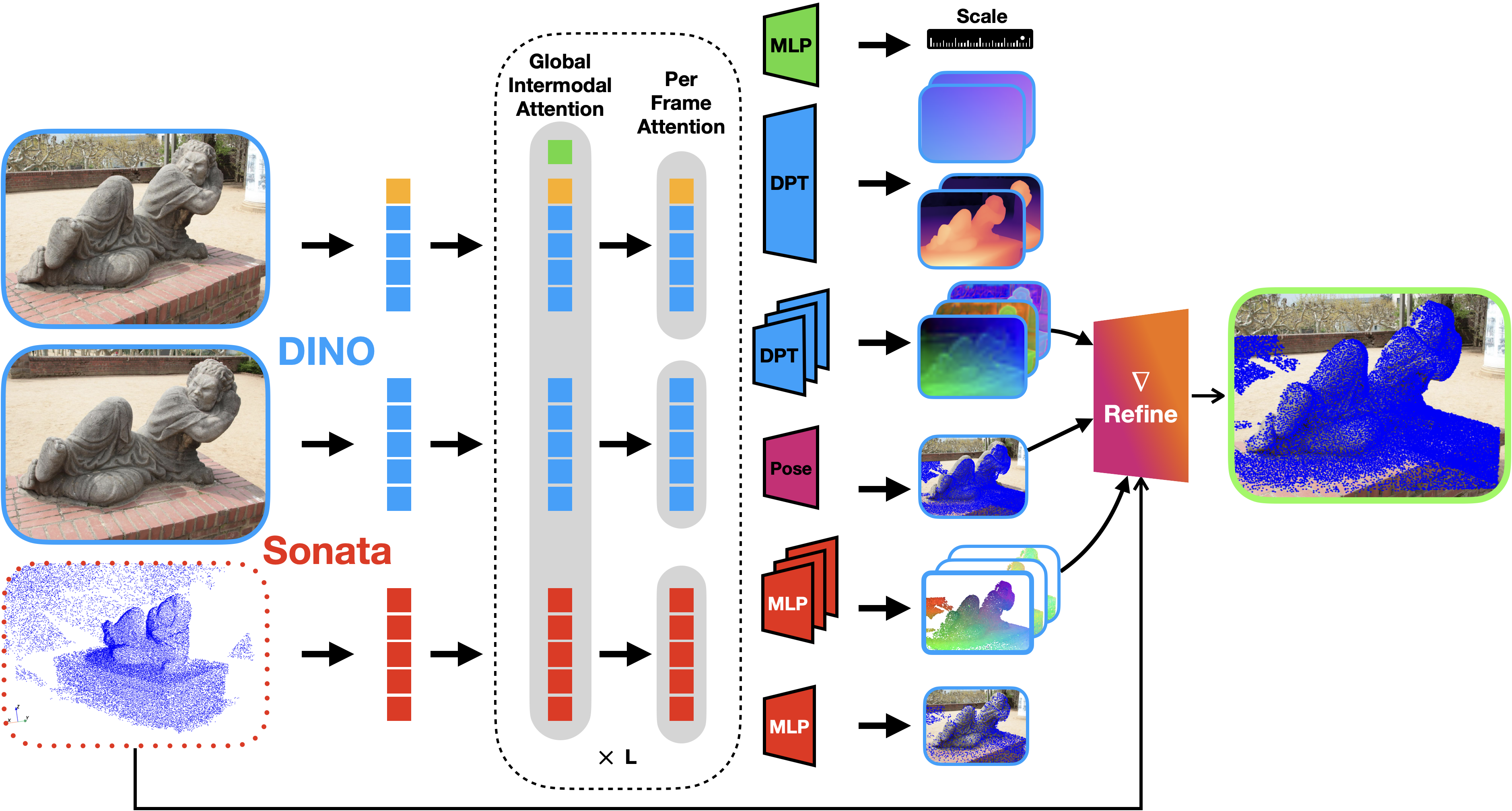}
    \caption{\textbf{DRS-VPT architecture.}
Query images and a reference 3D point cloud are encoded using DINOv2 and a Sonata point transformer respectively. Their tokens are fused using a transformer with alternating global cross-modal attention and modality-specific local attention. From the fused representation the model predicts camera poses, dense point maps, and alignment features for both modalities. The predicted scan pose provides an initial alignment which is refined using differentiable direct image–scan alignment, producing the final cloud pose in camera 1 frame.}
    \vspace{-15pt}
    \label{fig:architecture}
\end{figure}

\section{DRS-VPT}

Our end-to-end model (see~\autoref{fig:architecture}) ingests a set of $N$ RGB images together with a reference 3D point cloud.
We denote the image set as $\hat{\mathcal{I}} = \{ \hat{I}_i \}_{i=1}^{N}$ and the reference point cloud as $\hat{P}$.
All predicted geometric quantities are expressed in the coordinate frame of the first image $\hat{I}_1$, which serves as the reference frame for the scene.
During both training and inference the model operates on a small set of image observations (typically up to four views) jointly with the reference point cloud.
Crucially, the reference point cloud is not assumed to be aligned with any of the input images.
As a result, the model must infer the alignment between the visual observations and the 3D geometry of the scan while estimating the camera poses of the remaining images relative to $\hat{I}_1$.

Formally, the network implements

\begin{equation}
\begin{aligned}
f_{\text{DRS-VPT}}\big(\hat{\mathcal{I}}, \hat{P}\big) =
\Big\{ \,
&s,\,
(Q_i, \tilde{\mathbf{t}}_i, R_i, \tilde{D}_i, W_i^{X})_{i=1}^{N}, \\
&(\tilde{X}^{P}, Q^{P}, \tilde{\mathbf{t}}^{P}, W^{P}), \\
&(F_I^{(l)}, F_P^{(l)}, W_I^{(l)}, W_P^{(l)})_{l=1}^{3}
\Big\},
\end{aligned}
\end{equation}

Here $s \in \mathbb{R}^{+}$ denotes the predicted global metric scale.
For each image $i$, the model predicts a unit quaternion
$Q_i \in \mathbb{H}$ representing camera rotation and an up-to-scale translation
$\tilde{\mathbf{t}}_i \in \mathbb{R}^{3}$ expressed in the coordinate frame of the first image $\hat{I}_1$.
The tilde notation $(\tilde{\cdot})$ indicates quantities defined up to scale.

In addition, the model predicts per-view ray directions
$R_i \in \mathbb{R}^{3 \times H \times W}$ and up-to-scale ray depths
$\tilde{D}_i \in \mathbb{R}^{1 \times H \times W}$,
from which local point maps may be recovered.
For each reconstructed image point, the model also predicts confidence weights
$W_i^{X} \in \mathbb{R}^{H \times W}$ that estimate the reliability of the predicted geometry.

For the reference scan, the model predicts transformed scan points
$\tilde{X}^{P} \in \mathbb{R}^{3 \times K}$ together with a quaternion rotation
$Q^{P} \in \mathbb{H}$ and an up-to-scale translation
$\tilde{\mathbf{t}}^{P} \in \mathbb{R}^{3}$ describing the pose of the scan relative to the first camera $\hat{I}_1$.
The model additionally predicts per-point confidence weights
$W^{P} \in \mathbb{R}^{K}$ that capture the reliability of each transformed scan point.
The predicted scan pose provides an initialization for subsequent direct image–scan alignment.

To refine the scan's pose, the network also predicts dense features for both modalities.
We produce a three-level pyramid of coarse-to-fine alignment features
$\{F_I^{(l)}\}_{l=1}^{3}$ and $\{F_P^{(l)}\}_{l=1}^{3}$ for the first image and point cloud branches respectively,
alongside corresponding feature confidence maps
$\{W_I^{(l)}\}_{l=1}^{3}$ and $\{W_P^{(l)}\}_{l=1}^{3}$.
These weighted feature pyramids are used for differentiable direct alignment between the reference image and the 3D scan.

\subsection{Encoding Images and Point Clouds}
Given a set of $N$ images and a reference point cloud, DRS-VPT encodes both modalities into a shared latent space of dimension 1024.
For image inputs, we use the same frozen DINOv2 encoder as MapAnything~\cite{mapanything}.
Specifically, we extract the final-layer patch features from DINOv2 ViT-L, producing
\begin{equation}
    F_I \in \mathbb{R}^{1024 \times H/14 \times W/14},
\end{equation}
which serve as the per-patch image tokens used throughout the model.

For the reference point cloud, we employ the Sonata variant of PointTransformerV3~\cite{sonata, Wu2023PointTV}.
We use adaptive voxel downsampling with grid sizes ranging from \SI{5}{mm} to \SI{10}{cm} depending on the dataset and scene extent.
For non-metric datasets such as MegaDepth, the grid size is determined using the median nearest-neighbour distance of the point cloud.
To accommodate scenes of varying scale, the downsampled point cloud $\hat{P} \in \mathbb{R}^{3 \times K}$ is normalized by the scene scale (logarithm of the mean point range).
Sonata is adapted to produce features in the same latent dimension (1024).
The resulting encoder produces a set of per-point features
\begin{equation}
    F_P = \{ f^{\text{pt}}_k \}_{k=1}^{K}, \quad f^{\text{pt}}_k \in \mathbb{R}^{1024}.
\end{equation}
To help regress the original scene scale, the normalization term is encoded and added to the scan tokens via a learned gate initialised to zero.

\subsection{Cross-modal Attention}

The encoded tokens are processed by a transformer composed of alternating global and local attention layers.
An extra learnable scale token participates in global attention to propagate metric information across the image and point cloud representations.
A single \emph{base-frame embedding} is added to the tokens of the first image $\hat{I}_1$, defining the reference coordinate frame and anchoring the representation for subsequent geometric prediction.
We adopt a 24-layer alternating-attention architecture similar to MapAnything~\cite{mapanything}, consisting of 12 global and 12 local attention blocks with a latent dimension of 768 (projected down from 1024 at input), 12 attention heads, and an MLP ratio of 4.
The transformer weights are initialized from the corresponding MapAnything model and subsequently fine-tuned for the cross-modal registration task.

To maintain tractable attention complexity, the encoded point cloud tokens are uniformly downsampled to 5{,}000 tokens in Sonata's space-filling curve order before entering the transformer.
During the global attention blocks, all tokens from both modalities and the scale token attend jointly.
In the alternating local attention blocks, the modalities are processed independently. Point cloud tokens perform self-attention within the cloud branch and image tokens perform self-attention within each image view independently.

After the transformer layers, we fuse the output cloud tokens with their respective original high-resolution point features using a learned gating mechanism.
For each point, the final feature representation is computed as

\begin{equation}
    f_k^{\text{out}} =
\sigma(g_k) \, f_k^{\text{ctx}}
+
(1-\sigma(g_k)) \, f_k^{\text{local}},
\label{eq:upsample_cloud_token}
\end{equation}

where $f_k^{\text{ctx}}$ denotes the contextual feature obtained from the transformer tokens, $f_k^{\text{local}}$ is the original point encoder feature, and $\sigma(g_k)$ is a learned gating weight predicted from the concatenated features.
The gated fusion is specifically valuable because it recovers high-resolution per-point features that the transformer would otherwise lose due to the 5{,}000-token downsampling, with performance degrading when fewer points are available (see Tab.~\ref{tab:ablation-npoints}).

\subsection{Output Reconstruction}

For each image view $i$, the model predicts ray directions $R_i$, up-to-scale ray depths $\tilde{D}_i$, camera rotation $Q_i$, and up-to-scale translation $\tilde{\mathbf{t}}_i$, all expressed in the coordinate frame of the first image $\hat{I}_1$.
The image branch uses the same DPT-based depth and convolutional pose heads as MapAnything~\cite{mapanything}.
For the scan output, a lightweight linear decoder maps each upsampled cloud token (see~\autoref{eq:upsample_cloud_token}) to transformed coordinates $\tilde{X}^{P}$ in the frame of $\hat{I}_1$, alongside per-point confidence weights.
A global pose head additionally predicts scan rotation $Q^{P}$ and up-to-scale translation $\tilde{\mathbf{t}}^{P}$, providing a coarse initialization for the direct image--scan alignment stage.
The global metric scale $s$ is predicted using the scale token via a two-layer MLP with exponential output to ensure positivity.

From the first image and point cloud tokens, we decode three-level coarse-to-fine feature pyramids for cross-modal alignment.
Image features (dimension 24 at each level) are decoded using DPT heads with upsampling skip connections; point cloud features are decoded using a lightweight MLP with skip connections.
Both branches also output per-element confidence weights.
These pyramids drive the differentiable direct image--scan alignment described next.

\begin{figure}[t]
    \centering
    \includegraphics[width=\linewidth]{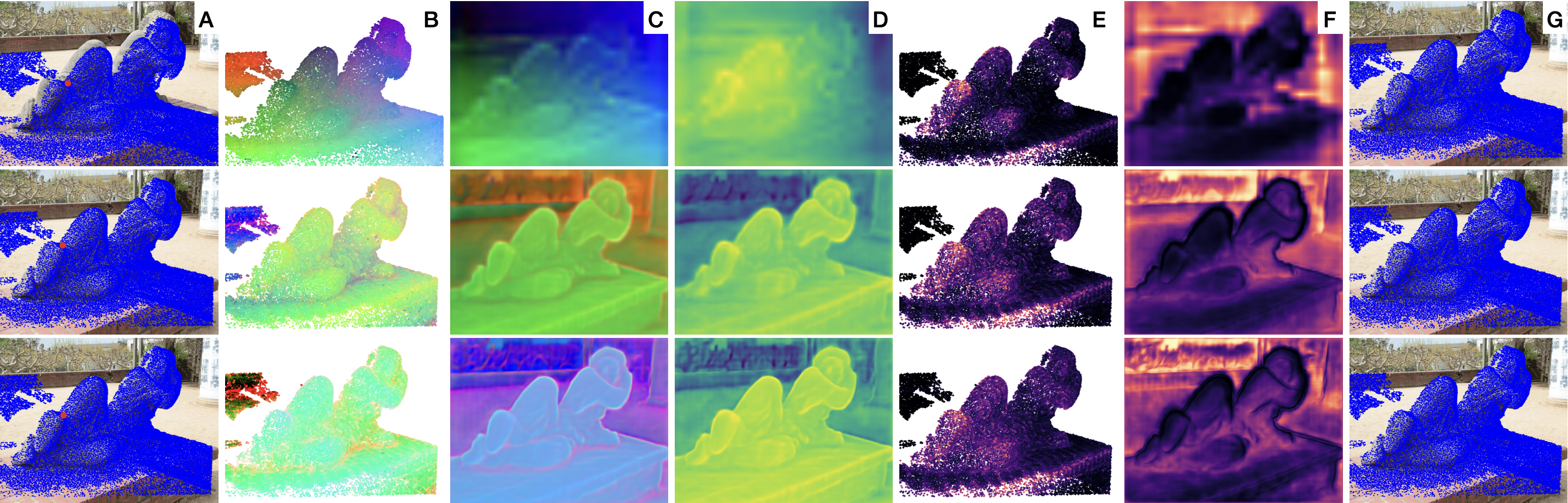}
    \caption{Coarse-to-fine Direct Alignment. The cloud pose estimate (A) is refined by aligning the cloud dense features (B) to the image features (C).
    (A) in the first row shows the pose head output for the cloud pose in camera 1 frame.
    Notice that it is slightly misaligned.
    The heatmap in (D) shows the cosine similarity of the queried cloud point marked in red in (A) to the feature at each pixel in (C).
    Note that this heatmap evolves from global saliency at the coarsest level (first row), to sharper, more local saliency at finer levels.
    (E) and (F) show the feature weights of the cloud and image features respectively. The post-alignment overlay for each pyramid level is shown in (G).
    See also~\autoref{fig:coarse_to_fine} for a more detailed example on ETH3D.}
    \label{fig:direct_alignment}
\end{figure}

\subsection{Unrolled Direct Image--Scan Alignment Refinement}
\label{sec:direct_alignment}

While the network predicts a decent initial pose for the reference scan, accurate cross-modal localization benefits from fine direct alignment as illustrated by the coarse-to-fine optimization process in Fig.~\ref{fig:direct_alignment}.

Let $\hat{\mathbf{P}}_j$ denote the input 3D measurements from the reference scan.
These points are not assumed to be aligned with the camera coordinate frame and remain in their original sensor coordinates.
Instead, the predicted global scale $s$ is used to rescale the translation component of the predicted scan pose $(Q^{P}, \tilde{\mathbf{t}}^{P})$, yielding the initialization $\mathbf{t}^{P} = s\,\tilde{\mathbf{t}}^{P}$ while the rotation is given by $R(Q^{P}) \in SO(3)$.
This scaled pose provides the initial alignment of the scan relative to the first camera view.

\begin{figure*}[t]
    \centering
    \includegraphics[width=\linewidth]{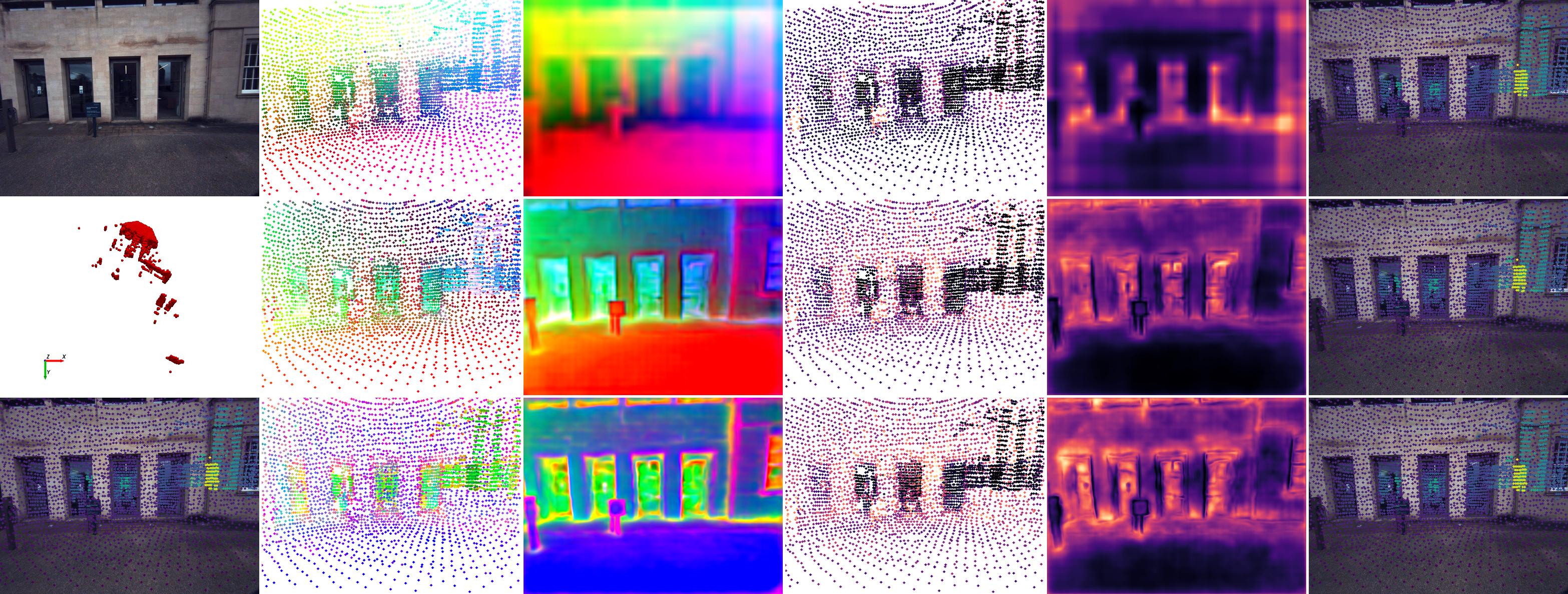}
    \caption{Learned feature saliency and confidence weighting on Oxford Spires. The model assigns high saliency to geometrically distinctive structures while the learned confidence weights (columns 4) suppress scan points that are occluded from the current camera viewpoint (pillars behind the facade of the building shown as dark streaks).}
    \label{fig:saliency}
    \vspace{-10pt}
\end{figure*}

Since the predicted scan pose is expressed in camera~1 coordinates, each scan point $\hat{\mathbf{P}}_j$ is projected into the image of camera~1 via the standard projection $\mathbf{u}_{ij} = \Pi\!\left(R(Q^{P})\,\hat{\mathbf{P}}_j + \mathbf{t}^{P}\right)$, given known intrinsics.
Following PixLoc~\cite{sarlin21pixloc}, at pyramid level $p$ we compute a feature residual between the point-cloud feature and the image feature bilinearly sampled at the projected pixel location, $\mathbf{r}^{p}_{j} = \mathbf{F}^{p}_{P,j} - \mathbf{F}^{p}_{I}\!\left(\mathbf{u}_{ij}\right) \in \mathbb{R}^{D_p}$, where $\mathbf{F}^{p}_{P,j}$ is the feature of point $j$ in the point-cloud branch and $\mathbf{F}^{p}_{I}(\cdot)$ is the image feature map evaluated at the projected location.
The alignment energy at level $p$ is then $E_p = \sum_{j} w^{P}_{j}\,w^{I}_{j}\,\rho\!\left(\|\mathbf{r}^{p}_{j}\|^2\right)$, where $w^{P}_{j}$ and $w^{I}_{j}$ are learned confidence weights for the point cloud and image features respectively, and $\rho(\cdot)$ is a robust penalty to suppress outliers.
The product of confidence weights allows the optimizer to jointly downweight unreliable geometric observations and ambiguous image regions.
Remarkably, the point-cloud weights $w^P_j$ often learn to downweight geometrically occluded points, such as scan measurements lying behind visible surfaces from the current camera viewpoint (see~\autoref{fig:saliency}).

To optimize the pose, we linearize the residuals with respect to an incremental pose update $\boldsymbol{\xi} \in \mathbb{R}^{6}$ in the Lie algebra of $\mathrm{SE}(3)$.
Stacking the residuals across all points and feature channels yields the Jacobian $ J_{k + (j-1)D_p} = \frac{\partial r^{p}_{j,k}}{\partial \boldsymbol{\xi}}$, where $r^{p}_{j,k}$ denotes the $k$-th feature channel of the residual $\mathbf{r}^{p}_{j}$.
Let $W$ denote a diagonal weight matrix containing the combined confidence weights and robust kernel derivatives.
The Gauss--Newton approximation to the Hessian is $H = J^\top W J$, and the pose increment is obtained by solving the normal equations,
\begin{equation}
    H\,\boldsymbol{\xi} = -J^\top W \mathbf{r},
\end{equation}
and applying the update via the $\mathrm{SE}(3)$ exponential map,
\begin{equation}
\begin{bmatrix}
R_{i+1} & \mathbf{t}_{i+1} \\
0 & 1
\end{bmatrix}
=
\exp(\widehat{\boldsymbol{\xi}})^\top
\begin{bmatrix}
R_i & \mathbf{t}_i \\
0 & 1
\end{bmatrix}.
\end{equation}

We unroll a fixed number of Gauss--Newton iterations (typically five) at each pyramid level and proceed from coarse to fine across the feature hierarchy.
The final supervision is applied using a reprojection loss between the predicted and ground-truth poses,

\begin{equation}
\mathcal{L}_{\text{reproj}}
=
\sum_{p}\sum_{j}
\rho\!\left(
\left\|
\Pi^{p}\!\left(\bar{R}\,\hat{\mathbf{P}}_{j} + \bar{\mathbf{t}}\right)
-
\Pi^{p}\!\left(R^{p}_{M}\,\hat{\mathbf{P}}_{j} + \mathbf{t}^{p}_{M}\right)
\right\|_2^2
\right),
\end{equation}

where $(\bar{R},\bar{\mathbf{t}})$ denotes the ground-truth pose and $(R^{p}_{M},\mathbf{t}^{p}_{M})$ denotes the pose after the final unrolled optimization step at pyramid level $p$.
This coarse-to-fine optimization allows the model to refine the initial scan alignment and achieve accurate image-to-scan localization.

\begin{figure*}[t]
    \centering
    \includegraphics[width=\linewidth]{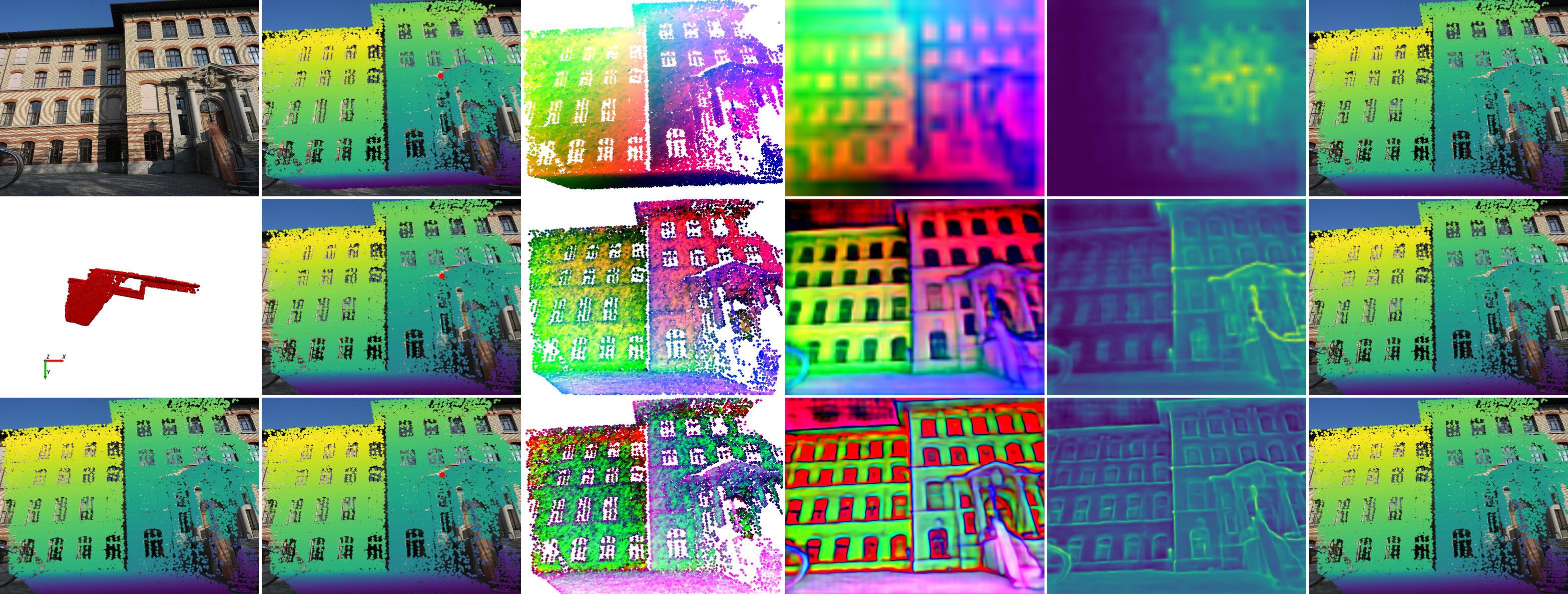}
    \caption{\textbf{Coarse-to-fine direct alignment on ETH3D Facade.} The first column shows the model's input image, the input misaligned cloud, and the final aligned overlay.
    Each subsequent column corresponds to one pyramid level, showing (top to bottom): the cloud overlay at the current estimate with a query point in red, point cloud features, image features, the query point's cross-modal similarity heatmap, and the overlay after refinement. The starting pose at the first level (top row) is produced by the transformer's scan pose head.}
    \label{fig:coarse_to_fine}
    \vspace{-10pt}
\end{figure*}

\subsection{Training Loss Formulation}


We supervise the predicted ray directions $R_i$, rotations $Q_i$, translations $\tilde{\mathbf{t}}_i$, depths $\tilde{D}_i$, local point maps $\tilde{L}_i$, world-frame point maps $\tilde{X}_i$, and confidence maps $C_i$ against their ground-truth counterparts $\hat{R}_i$, $\hat{Q}_i$, $\hat{\mathbf{t}}_i$, $\hat{D}_i$, $\hat{L}_i$, and $\hat{X}_i$.

The ray and rotation losses are computed using $\ell_2$ distances,

$\mathcal{L}_{\text{rays}} = \sum_i \|\hat{R}_i - R_i\|_2$ and
$\mathcal{L}_{\text{rot}} = \sum_i \min(\|\hat{Q}_i - Q_i\|_2, \|-\hat{Q}_i - Q_i\|_2)$,
which accounts for the two-to-one symmetry of unit quaternions.
Translations are optimized in a scale-invariant manner,
$\mathcal{L}_{\text{translation}} =
\sum_i \left\|\frac{\hat{\mathbf{t}}_i}{\hat{z}} -
\frac{\tilde{\mathbf{t}}_i}{\tilde{z}}\right\|_2$,
where $\hat{z}$ and $\tilde{z}$ denote the ground-truth and predicted scene scales.

We train depth and point-map predictions in log-space using, \\ $f_{\log}(x)=\frac{x}{\|x\|}\log(1+\|x\|)$.
The corresponding losses are \\
$\mathcal{L}_{\text{depth}}=\sum_i\|f_{\log}(\hat{D}_i/\hat{z})-f_{\log}(\tilde{D}_i/\tilde{z})\|_2$,
$\mathcal{L}_{\text{lpm}}=\sum_i\|f_{\log}(\hat{L}_i/\hat{z})-f_{\log}(\tilde{L}_i/\tilde{z})\|_2$,
and the confidence-weighted point-map loss, \\
$\mathcal{L}_{\text{pointmap}}=\sum_i C_i\|f_{\log}(\hat{X}_i/\hat{z})-f_{\log}(\tilde{X}_i/\tilde{z})\|_2-\alpha\log C_i$.

The global metric scale $s$ is optimized using
$\mathcal{L}_{\text{scale}}=\|f_{\log}(\hat{z})-f_{\log}(z_{\text{metric}})\|_2$,
where $z_{\text{metric}} = s\cdot \mathrm{sg}(\tilde{z})$,
combines the predicted scale $s$ with a detached normalization term $\tilde{z}$ to prevent gradient coupling through scale.

Confidence-weighted point-map supervision is applied to both image and point cloud outputs.
Scale supervision is included only for datasets with metric ground truth.

The final training objective combines geometric supervision with the reprojection loss from the differentiable alignment stage,

\begin{align}
\mathcal{L}_{\text{total}} =
& 10\,\mathcal{L}_{\text{pointmap}} +
\mathcal{L}_{\text{rays}} +
\mathcal{L}_{\text{rot}} +
\mathcal{L}_{\text{translation}} \nonumber \\
& +
\mathcal{L}_{\text{depth}} +
\mathcal{L}_{\text{lpm}} +
\mathcal{L}_{\text{scale}} +
\mathcal{L}_{\text{reproj}} .
\end{align}

\subsection{Datasets and Multi-View Sampling}

We train DRS-VPT on a diverse collection of multi-modal datasets spanning indoor, outdoor, and large-scale driving environments:
MegaDepth~\cite{li2018megadepth}, OxfordSpires~\cite{tao2025spires}, KITTI~\cite{geiger2012kitti}, ETH3D~\cite{schops2017eth3d}, Argoverse~\cite{argoverse}, nuScenes~\cite{nuscenes}, PandaSet~\cite{pandaset}, 7-Scenes~\cite{7scenes}, ScanNet++~\cite{scannetpp}, WildRGBD~\cite{wildrgbd}, and GrandTour~\cite{boxi,grand_tour}.
These datasets provide a wide range of sensing modalities including structure-from-motion reconstructions, LiDAR scans, and dense laser-scanner depth maps.
Training across this heterogeneous collection enables the model to generalize across both metric and up-to-scale geometric regimes as well as across different sensor configurations.

\paragraph{Datasets.}
For \textbf{MegaDepth}, we use the publicly available structure-from-motion reconstructions, treating the recovered 3D points as up-to-scale supervision for the point cloud branch.
\textbf{OxfordSpires} provides synchronized camera–LiDAR data collected in outdoor environments; we use sparse LiDAR returns within the camera frustums for geometric supervision.
In \textbf{KITTI}, \textbf{Argoverse}, \textbf{nuScenes}, and \textbf{PandaSet}, the model ingests full LiDAR scans captured by autonomous driving platforms,
providing the low-overlap alignment challenge between narrow-FOV cameras and full 360$^{\circ}$ vehicle-mounted LiDAR sweeps.
\textbf{ETH3D}, \textbf{ScanNet++}, and \textbf{7-Scenes}, \textbf{WildRGBD} provide dense indoor reconstructions obtained from structured-light, depth camera or laser scanning systems,
which we treat as depth-derived point clouds associated with the corresponding camera views.
Finally, the \textbf{GrandTour} dataset contributes camera--LiDAR sequences from a robot-mounted platform across deployment-relevant in-the-wild settings such as forests and warehouses, providing the model with registration experience in the unstructured environments where robotic systems must operate. 

\paragraph{Multi-View Sampling.}
To ensure consistent geometric overlap during training, we adopt a covisibility-based sampling strategy similar to MapAnything~\cite{mapanything}.
For each dataset, we precompute pairwise frame overlaps using reprojection-based visibility checks derived from ground-truth poses or depth maps.
Training samples are constructed such that all selected image views share at least 25\% geometric overlap with the reference scan.

During training, we sample connected sets of $N$ covisible images together with a reference point cloud.
The reference scan may correspond either to a LiDAR frame or to a reconstructed point cloud associated with one of the views.
To encourage generalizability, we apply random rotation and translation augmentations to the input scan during training.

For datasets with smaller-scale scenes (e.g., WildRGBD, and ScanNet++), we occasionally aggregate several nearby point clouds into a single reference frame before presenting them to the model.
This produces larger map-like scans while preserving the single-scan input formulation of the network.
For 7-Scenes, all posed depths are aggregated into a map frame and downsampled to train relocalization.
Consequently, the model is consistently trained in an image-to-scan relocalization setting rather than in a local reconstruction configuration.


\section{Experiments}

\begin{table}[t]
\caption{
Camera--LiDAR alignment under wide initialization.
For each dataset we report translation $t$ (in meters) and rotation $r$
(in degrees) errors, shown as mean followed by median.
Accuracy is the percentage of predictions within 2\,m translation and
5$^\circ$ rotation error.
}
\label{tab:wide-kitti-nuscenes}
\centering
\scriptsize
\renewcommand{\arraystretch}{1.05}
\setlength{\tabcolsep}{3.0pt}

\begin{tabular}{@{}l ccc @{\hspace{6pt}} ccc}
\toprule
& \multicolumn{3}{c}{\textbf{KITTI}}
& \multicolumn{3}{c}{\textbf{nuScenes}} \\
\cmidrule(lr){2-4} \cmidrule(lr){5-7}
\textbf{Method}
& $t$ [m] & $r$ [$^\circ$] & Acc
& $t$ [m] & $r$ [$^\circ$] & Acc \\
\midrule

CorrI2P~\cite{corri2p}
& 3.78 / -- & 5.89 / -- & 72.42
& 3.04 / -- & 3.73 / -- & 49.00 \\

VP2P~\cite{vp2p}
& 1.91 / 1.23 & 5.17 / 2.69 & 64.13
& 0.89 / -- & 2.15 / -- & 88.33 \\

CoFiI2P~\cite{cofii2p}
& 0.32 / 0.27 & 1.27 / 1.06 & 99.41
& 1.21 / 0.75 & 2.54 / 1.70 & 90.87 \\

TrafficLoc~\cite{trafficloc}
& 0.17 / 0.14 & 0.82 / 0.55 & 99.18
& 2.26 / 0.23 & 5.68 / 1.63 & 90.06 \\

\midrule
\textbf{DRS-VPT (Ours)}
& \textbf{0.09} / \textbf{0.05} & \textbf{0.34} / \textbf{0.25} & \textbf{99.73}
& \textbf{0.33} / \textbf{0.13} & \textbf{0.67} / \textbf{0.33} & \textbf{97.76} \\

\bottomrule
\end{tabular}
\end{table}

We evaluate DRS-VPT on the task of image-to-scan relocalization across both outdoor autonomous driving and indoor RGB-D environments.
Unlike most prior relocalization approaches, which train scene-specific
models tied to a fixed map representation, DRS-VPT directly ingests a
3D scan and image at inference time and predicts camera poses relative to that scan.

We evaluate the method in three complementary settings.
First, we measure wide-baseline camera--LiDAR alignment accuracy on two autonomous driving datasets --- KITTI and nuScenes.
Second, we evaluate indoor camera relocalization on the 7Scenes benchmark, where DRS-VPT dynamically ingests the scene map instead of learning a dedicated scene model.
Finally, we test cross-dataset generalization by localizing images in completely unseen environments without fine-tuning.

Together, these experiments evaluate in-domain performance as well as the ability of a unified multi-modal transformer to generalize across different sensors, scene scales, and environmental statistics.

\subsection{Autonomous Driving Alignment}

We first evaluate wide-baseline camera--LiDAR alignment on KITTI and nuScenes.
Following prior work, we report translation and rotation errors together with the percentage of predictions within 2\,m translation and 5$^\circ$ rotation thresholds.

For this experiment, we use KITTI odometry sequences~9 and~10 for the wide-baseline localization task.
None of these sequences are included in our training set, making this a genuine held-out evaluation.

In the wide-baseline setting, we perturb the LiDAR point cloud with a random rigid transformation of up to $\pm 10$\,m in the $x$--$z$ plane and an arbitrary rotation about the vertical ($y$) axis, following the standard evaluation protocol of prior camera--LiDAR registration methods~\cite{corri2p,vp2p}. This yaw-and-translation protocol reflects the dominant degrees of freedom in ground-vehicle operation; the model receives no initial pose and must recover the full alignment globally from the perturbed input.

Table~\ref{tab:wide-kitti-nuscenes} shows that DRS-VPT achieves state-of-the-art performance on both datasets.
Despite operating with a unified architecture that directly processes point clouds and images, our method outperforms specialized correspondence-based methods such as CorrI2P, VP2P, CoFiI2P, and TrafficLoc.
Notably, the model achieves the lowest translation and rotation errors on KITTI while maintaining strong performance on nuScenes, demonstrating that the learned geometric representation transfers effectively across different autonomous driving sensor configurations.

\subsection{Indoor Camera--Map Localization}

We evaluate indoor relocalization on the 7Scenes dataset, a standard benchmark for camera pose estimation in RGB-D environments.
Unlike most existing methods, which train scene-specific neural models that encode a map representation during training, DRS-VPT directly ingests a 3D geometric map at inference time and predicts the camera pose relative to that map.

For this experiment, 3D maps are constructed from the RGB-D sequences provided with the dataset.
3D maps are constructed from the training sequences of each scene; at evaluation time, query images are drawn from the held-out test split and localized against these training-sequence maps.

Table~\ref{tab:7scenes_methods_rows} compares our approach to several established relocalization systems.
Methods such as DSAC* and ACE learn scene-specific neural scene representations, while PixLoc performs feature-based alignment between query images and reference images stored in a map.
In all cases, these baselines operate entirely within the image domain, matching pixels between query and reference views captured by the same camera and relying on scene-specific training to encode the map structure.

In contrast, DRS-VPT localizes query images directly against a 3D map constructed from depth observations, introducing a cross-modal alignment problem between RGB images and point clouds.
Despite this additional modality gap and the absence of scene-specific training, our method remains competitive with existing approaches, demonstrating that the learned cross-modal features provide sufficient geometric signal for reliable localization.

Performance on the \textit{Stairs} scene is noticeably lower than on other scenes.
This environment contains highly repetitive staircase structures that produce strong geometric aliasing, making pose disambiguation difficult even for specialized relocalization systems.

These results suggest that dynamic map ingestion may provide a viable alternative to scene-specific relocalization models.
We further validate this hypothesis in the next experiment, where we localize images in a completely unseen indoor dataset using only the provided 3D map, demonstrating strong cross-environment generalization.

\begin{table}[t]
\centering
\footnotesize
\caption{Pose estimation results on the 7Scenes dataset (indoor).
Values are translation error (cm) / rotation error (deg).
The last column reports overall recall (\%).}
\label{tab:7scenes_methods_rows}

\setlength{\tabcolsep}{1.5pt} 
\begin{tabular}{@{}lcccccccc@{}}
\toprule
Method
& Chess & Fire & Heads & Office & Pumpkin & Kitchen & Stairs & Recall$\uparrow$ \\
\midrule

InLoc \cite{taira2018inloc}
& 3/1.05 & 3/1.07 & 2/1.16 & 3/1.05 & 5/1.55 & 4/1.31 & 9/2.47 & 66.3 \\

DSAC* \cite{brachmann2021dsacstar}
& \textbf{2}/1.10 & \textbf{2}/1.24 & \textbf{1}/1.82 & 3/1.15 & \textbf{4}/1.34 & 4/1.68 & \textbf{3}/\textbf{1.16} & \textbf{85.2} \\


PixLoc \cite{sarlin21pixloc}
& \textbf{2}/0.80 & \textbf{2}/\textbf{0.73} & \textbf{1}/0.82 & 3/\textbf{0.82} & \textbf{4}/1.21 & \textbf{3}/\textbf{1.20} & 5/1.30 & 75.7 \\

ACE \cite{brachmann2023ace}
& \textbf{2}/\textbf{0.68} & \textbf{2}/0.87 & \textbf{1}/\textbf{0.67} & 3/0.84 & \textbf{4}/\textbf{1.15} & 4/1.34 & 5/1.25 & 75.6 \\

\midrule
\textbf{DRS-VPT (ours)}
& 3/0.84 & \textbf{2}/1.01 & 2/1.00 & 3/0.88 & 7/1.67 & 5/1.30 & 27/3.26 & 66.4 \\

\bottomrule
\end{tabular}
\end{table}

\subsection{Zero-Shot Indoor Relocalization}

To further evaluate cross-environment generalization, we test DRS-VPT on the
12Scenes\cite{12scenes} dataset, which contains indoor environments similar to 7Scenes but
was never used during training. As in the previous experiment, the model
localizes query images against 3D maps constructed from depth-camera
observations of each scene.

Table~\ref{tab:twelve_scenes_scene_agg} reports aggregated results across
subscenes, showing median translation and rotation errors together with
recall under the standard $5$\,cm and $5^\circ$ accuracy threshold. Despite
never observing these environments during training, DRS-VPT achieves
consistent localization across all scenes with an overall recall of
53.6\%.

Although recall is lower than on 7Scenes due to the domain shift and the
absence of scene-specific training, the results demonstrate that the model
can successfully localize images in completely unseen environments using
only a provided geometric map. This experiment highlights the ability of
DRS-VPT to generalize beyond the training distribution and operate in a
true zero-shot relocalization setting.

\begin{table}[t]
\centering
\scriptsize
\setlength{\tabcolsep}{4pt}
\renewcommand{\arraystretch}{1.08}

\begin{minipage}{0.48\linewidth}
\centering
\caption{DRS-VPT results on Twelve Scenes (scene-level aggregation).}
\label{tab:twelve_scenes_scene_agg}
\begin{tabular}{lcccc}
\toprule
 & \textbf{apt1} & \textbf{apt2} & \textbf{office1} & \textbf{office2} \\
\midrule
Trans (cm)   & 4    & 5    & 4    & 7    \\
Rot (deg)    & 1.92 & 2.23 & 1.82 & 2.78 \\
Recall (\%)  & 56.8 & 48.9 & 62.0 & 33.7 \\
\bottomrule
\end{tabular}
\end{minipage}
\hfill
\begin{minipage}{0.48\linewidth}
\centering
\caption{KITTI $\rightarrow$ nuScenes generalization for image-to-point alignment.}
\label{tab:cross_i2p_kitti_to_nuscenes}
\begin{tabular}{lccc}
\toprule
\textbf{Method} & $\mathbf{t\,[m]}$ & $\mathbf{r\,[^\circ]}$ & \textbf{Recall (\%)} \\
\midrule
TrafficLoc & 16.55 & 95.55 & 1.9 \\
CoFiI2P    & 17.46 & 93.40 & 2.9 \\
\textbf{DRS-VPT} & \textbf{0.538} & \textbf{1.14} & \textbf{66.74} \\
\bottomrule
\end{tabular}
\end{minipage}
\vspace{-15pt}
\end{table}

\subsection{Cross-Dataset Generalization}

Finally, we evaluate cross-dataset generalization in the autonomous driving setting.
For this experiment, baseline methods are trained on KITTI and evaluated directly on nuScenes, following the protocol used in prior work.
For DRS-VPT, we train a single model on our full multi-dataset training corpus while explicitly excluding nuScenes.

Table~\ref{tab:cross_i2p_kitti_to_nuscenes} reports localization performance on nuScenes.
Methods trained only on KITTI fail to generalize under the significant domain shift between the two datasets, resulting in extremely large pose errors and near-zero recall.
In contrast, DRS-VPT maintains accurate alignment with a median translation error of 0.538\,m and rotation error of $1.14^\circ$, achieving a recall of 66.74\%.

These results demonstrate that the geometric representations learned by DRS-VPT transfer effectively across different autonomous driving platforms and sensor configurations.
More broadly, the ability to localize images in previously unseen environments using only a provided 3D map suggests that dynamic map ingestion can serve as a practical alternative to scene-specific localization pipelines.

\subsection{Ablation: Refinement Strategy and Point Density}

We also explored geometric alignment on predicted point maps as an alternative refinement strategy. Tab.~\ref{tab:ablation-refinement} compares Iterative Closest Point (ICP, aligning the camera-view point map to the input cloud) and Kabsch (aligning the input cloud to the cloud-head point map).
While these methods improve rotation, neither significantly reduces translation error.
We observe that point-based alignment demands globally accurate metric depth from the model---a fundamentally harder regression target than learning cross-modal feature alignment.
Feature-metric direct alignment instead exploits discontinuities in learned cross-modal features at surface boundaries,
a signal the end-to-end training naturally produces, reducing mean translation error by $5\times$.
Tab.~\ref{tab:ablation-npoints} further shows performance improving with point density, demonstrating the benefits of the gated fusion mechanism (see~\autoref{eq:upsample_cloud_token}) in recovering high-resolution per-point detail.

\vspace{-20pt}
\begin{figure}[h]
\begin{minipage}[c]{0.54\linewidth}
\captionof{table}{
Comparison of refinement strategies on KITTI.
}
\label{tab:ablation-refinement}
\centering
\scriptsize
\renewcommand{\arraystretch}{1.05}
\setlength{\tabcolsep}{1.5pt}
\begin{tabular}{@{}l ccc}
\toprule
\textbf{Method} & $t$ [m] & $r$ [$^\circ$] & Acc \\
\midrule
Raw pose head                        & 0.50 / 0.44 & 1.45 / 1.38 & 99.64 \\
+ ICP refinement                     & 0.52 / 0.45 & 1.25 / 1.12 & 99.57 \\
+ Kabsch alignment                   & 0.61 / 0.52 & 0.94 / 0.83 & 98.82 \\
\textbf{+ Feature-metric} & \textbf{0.09} / \textbf{0.05} & \textbf{0.34} / \textbf{0.25} & \textbf{99.73} \\
\bottomrule
\end{tabular}
\end{minipage}
\hfill
\begin{minipage}[c]{0.44\linewidth}
\captionof{table}{
Robustness of GN refinement to point cloud density on KITTI.
}
\label{tab:ablation-npoints}
\centering
\scriptsize
\renewcommand{\arraystretch}{1.05}
\setlength{\tabcolsep}{1.5pt}
\begin{tabular}{@{}l ccc}
\toprule
\textbf{N Points} & $t$ [m] & $r$ [$^\circ$] & Acc \\
\midrule
5{,}000  & 0.14 / 0.09 & 0.46 / 0.35 & 99.64 \\
10{,}000 & 0.10 / 0.06 & 0.35 / 0.29 & 99.71 \\
20{,}000 & 0.09 / 0.06 & 0.34 / 0.25 & 99.71 \\
\textbf{50{,}000} & \textbf{0.08} / \textbf{0.05} & \textbf{0.32} / \textbf{0.23} & \textbf{99.71} \\
\bottomrule
\end{tabular}
\end{minipage}
\end{figure}
\vspace{-20pt}



\section{Discussion and Future Work}

We presented DRS-VPT, a feed-forward transformer for image-to-scan relocalization that unifies geometric prediction and differentiable direct alignment in a single architecture.
Unlike prior methods that explicitly classify frustum overlap before solving for correspondences, DRS-VPT learns to identify relevant scan regions implicitly through the reprojective inductive bias of the alignment objective (see~\autoref{fig:task_overview}).
Strikingly, the learned confidence weights suppress back-facing, occluded scan points without any explicit occlusion supervision (see~\autoref{fig:direct_alignment}), evidence that the reprojective objective instils physically meaningful geometric understanding.
Across autonomous driving and indoor benchmarks, DRS-VPT achieves state-of-the-art wide-baseline camera--LiDAR alignment, remains competitive with specialized indoor relocalization systems, and generalizes well to unseen environments. 

Several open directions remain.
The current formulation assumes a single rigid reference scan, therefore, extending to multi-scan or hierarchical map representations would broaden applicability to large-scale environments.
Handling articulated and dynamic objects would open applications in robot state tracking.
Additionally, generalizing to allowing multiple output instances per point cloud, unlocks object-level instance detection as a downstream task.

\section*{Acknowledgements}
This work has been carried out within the framework of the EUROfusion Consortium, funded by the European Union via the Euratom Research and Training Programme (Grant Agreement No 101052200 — EUROfusion) and from the EPSRC [grant number EP/W006839/1. Views and opinions expressed are however those of the author(s) only and do not necessarily reflect those of the European Union or the European Commission. Neither the European Union nor the European Commission can be held responsible for them.
The authors thank Matias Mattamala, Edgar Sucar, Eldar Insafutdinov and Benjamin Ramtoula for valuable discussions on earlier versions of this work, and Emil Jonasson and Yifu Tao for their careful review of the manuscript.

%
%
\bibliographystyle{splncs04}
\bibliography{main}

\end{document}